\documentclass{article}
\usepackage{ijcai24}

\usepackage{times}
\usepackage{soul}
\usepackage{url}
\usepackage[hidelinks]{hyperref}
\usepackage[utf8]{inputenc}
\usepackage[small]{caption}
\usepackage{graphicx}
\usepackage{amsmath}
\usepackage{amsthm}
\usepackage{booktabs}
\usepackage{algorithm}
\usepackage{algorithmicx}
\usepackage[switch]{lineno}

\usepackage{bbding}
\usepackage[table,xcdraw]{xcolor}
\usepackage{subcaption}
\usepackage{multirow}
\usepackage{amsfonts}       
\usepackage{nicefrac}       
\usepackage{microtype}      
\usepackage{setspace}
\usepackage{algpseudocode}
\usepackage{amssymb}

\title{BiTA:~Bi-Directional Tuning for Lossless Acceleration in Large Language Models}

\author{
Feng Lin$^{1,2}$
\and
Hanling Yi$^1$\and
Hongbin Li$^1$\and
Yifan Yang$^1$\and
Xiaotian YU$^1$\and\\
Guangming Lu$^2$\And
Rong Xiao$^1$\\
\affiliations
$^1$Intellifusion Inc.\\
$^2$Harbin Institute of Technology, Shenzhen\\
\emails
lin1993@mail.ustc.edu.cn,\\
\{hanling.cuhk, lee.blingner, yifan.yang.cn, xiaotianyu.ac, rongxiao\}@gmail.com,\\
luguangm@hit.edu.cn
}

\begin{document}

\maketitle

\begin{abstract}
    Large language models (LLMs) commonly employ autoregressive generation during inference, leading to high memory bandwidth demand and consequently extended latency. To mitigate this inefficiency, we present \textbf{Bi}-directional \textbf{T}uning for lossless \textbf{A}cceleration (BiTA), an innovative method expediting LLMs via streamlined semi-autoregressive generation and draft verification. Inspired by the concept of prompt tuning, we enhance LLMs with a parameter-efficient design called bi-directional tuning for the capability in semi-autoregressive generation. Employing efficient tree-based decoding, the models perform draft candidate generation and verification in parallel, ensuring outputs identical to their autoregressive counterparts under greedy sampling. BiTA serves as a lightweight plug-in module, seamlessly boosting the inference efficiency of existing LLMs without requiring additional assistance models or incurring significant extra memory costs. Applying the proposed BiTA, LLaMA-2-70B-Chat achieves a $2.7\times$ speedup on the MT-Bench benchmark. Extensive experiments confirm our method surpasses state-of-the-art acceleration techniques.
\end{abstract}

\section{Introduction}
Recent years have witnessed a rapid evolution in large language models (LLMs) grounded in transformer architectures. The parameters of LLMs have swiftly burgeoned, spanning from several billions to tens of trillions, as exemplified by models like Chat-GPT~\cite{brown2020language}, LLaMA-2~\cite{touvron2023llama2}, and others. While LLMs exhibit diverse and powerful generative capabilities, they encounter challenges in inference latency due to the substantial computational burden arising from numerous parameters. As a result, accelerating the inference process of LLMs has become a significant focus, particularly in resource-limited scenarios such as edge devices and real-time applications like chatbots.

The prevalent decoder-only LLMs, highlighted in recent works~\cite{zhang2022opt,workshop2022bloom,almazrouei2023falcon}, adhere to a token-by-token generation manner. Each token generated necessitates a distinct inference execution, reflecting their autoregressive (AR) generation nature and leading to a substantial number of transformer calls during inference. These calls frequently encounter constraints associated with memory bandwidth, causing reduced computational efficiency and prolonged wall-clock times~\cite{hooper2023speed,zhang2023draft,shazeer2019fast}.  Thus, a key strategy to expedite LLMs is to minimize the number of inference executions.

\begin{figure}[t]
\centering
\includegraphics[width=3.3in]{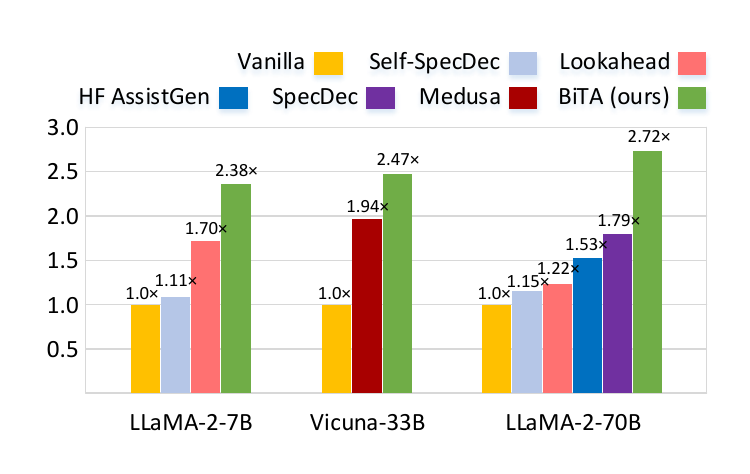}
\vspace{-5pt}
\caption{A comparison of LLM acceleration techniques, encompassing both state-of-the-art methods and our approach, is presented on MT-Bench using various base models. The speedup numbers are either sourced from the respective papers or reproduced using the released source codes in a standardized hardware environment by us, in cases where explicit disclosure is not provided.}
\label{fig:demo}
\end{figure}

Semi-autoregressive (SAR) decoding, as introduced in machine translation literature~\cite{wang2018semi}, mitigates the high demand for inference executions by producing multiple tokens in parallel with a single step of model inference. However, the majority of current LLMs are AR models, lacking the capability for SAR generation. Re-training for an SAR model appears challenging due to the misalignment between the SAR objectives and AR pretraining. Additionally, training from scratch in an SAR fashion seems almost impractical considering the substantial resource consumption involved. Furthermore, SAR models often suffer from quality degradation compared to their AR counterparts~\cite{wang2018semi,gu2020fully,huang2022directed}.

Is it feasible to empower an existing AR language model to function as an SAR one with minimal adaptation while ensuring satisfactory model performance? We break down the question into two aspects and address each separately. Leveraging advancements in parameter-efficient tuning (PET) techniques, especially prompt tuning~\cite{lester2021power}, we seamlessly enhance AR models with the capability of SAR generation. Simultaneously, drawing insights from speculative decoding~\cite{leviathan2023fast}, which typically follows the ``draft-then-verify'' paradigm, we perform verification on SAR outputs. This careful verification ensures that SAR outputs remain consistent when inferred in an AR manner, thereby preventing the typical degradation observed in standard SAR models.

In this paper, we propose \textbf{Bi}-directional \textbf{T}uning for lossless \textbf{A}cceleration (BiTA), a novel acceleration scheme aimed at achieving lossless SAR decoding for AR language models by learning very limited additional learnable parameters. Specifically, BiTA comprises two components: SAR draft generation through the proposed bi-directional tuning and streamlined verification for generated draft candidates. Bi-directional tuning involves incorporating both prompt tokens and mask tokens to enable the prediction of future tokens, extending beyond the next token for an AR model. This method is metaphorically referred to as the learnable prefix and suffix embeddings in token sequence. Through an elaborate tree-based attention mechanism, generation and verification operate simultaneously in a single forward pass within the converted AR model. The universal design eliminates the need for extra validation steps or external verification models. Benefiting from the concept of prompt tuning, the proposed method can function as a plug-and-play module for expediting any publicly available transformer-based LLMs, particularly those well-instructed chatbots~\cite{touvron2023llama2,vicuna2023,almazrouei2023falcon}, without compromising their strong generative capabilities. Our primary contributions can be summarized as follows:
\begin{itemize}
\item To reduce transformer calls in AR generation, we adapt AR language models for SAR generation using the proposed bi-directional tuning, introducing as few as $0.01\%$ additional trainable parameters.
\item We introduce an efficient tree-based decoding for SAR outputs, enabling simultaneous draft generation and verification, thereby eliminating the necessity for extra validation steps or external models.
\item BiTA operates as a plug-and-play module, applicable to any publicly available transformer-based LLMs without altering the original outputs. With the aid of BiTA, LLaMA-2-70B-Chat achieves a $2.7\times$ speedup on the MT-Bench benchmark, surpassing state-of-the-art techniques in extensive experiments (A brief comparison is presented in Figure~\ref{fig:demo}).
\end{itemize}

\section{Related Work}
\subsection{LLM Acceleration}
LLM acceleration can be approached through various dimensions, including model compression~\cite{hinton2015distilling,liu2018rethinking}, architecture simplification~\cite{dao2022flashattention}, quantization~\cite{gholami2022survey}, memory management~\cite{kwon2023efficient}, kernel optimization~\cite{wang2021lightseq}, inference scheduling~\cite{kwon2023efficient}, efficient decoding~\cite{santilli2023accelerating}, and more. These techniques span from cutting-edge algorithmic modifications to groundbreaking changes in system designs, finding widespread applications in practical scenarios~\cite{miao2023towards}. 

In this paper, a specific emphasis is placed on SAR decoding as one of the typical methods for efficient decoding. SAR decoding, derived from non-autoregressive (NAR) decoding~\cite{gu2018non}, is initially introduced for machine translation~\cite{stern2018blockwise}. It diverges from the conventional AR generation paradigm by decoding output tokens in parallel, with the goal of attaining AR output quality through post-processing strategies~\cite{xiao2023survey}.

\subsection{Speculative Decoding}
Speculative decoding stands out as another typical efficient decoding method, involving the anticipation of token distribution of corresponding AR models in a speculative manner. An early method~\cite{stern2018blockwise} generates future predictions as drafts by auxiliary prediction heads, then validate them by a scoring model. Recent studies~\cite{leviathan2023fast,chen2023accelerating} utilize external draft models for token distribution sampling from large target models. SpecDec~\cite{xia2023speculative} explores the designing principles for efficient draft models. OSD~\cite{liu2023online} enhances draft models through online retraining.

Without employing external draft models, SPEED~\cite{hooper2023speed} designs a faster speculative pipeline with cyclic parameter sharing. While Self-SpecDec~\cite{zhang2023draft} expedites drafting by selectively skipping specific intermediate layers. Medusa~\cite{medusa} adopts multiple additional prediction heads, akin to literature~\cite{stern2018blockwise}. PaSS~\cite{monea2023pass} obtains SAR drafts by means of ``look-ahead'' embeddings. REST~\cite{he2023rest} utilizes the knowledge retrieval. Lookahead~\cite{fu2023lookahead} relies solely on n-grams generated by LLMs as speculative draft candidates. Optimizing verification is another way. SpecInfer~\cite{miao2023specinfer} uses a draft candidate token tree for parallel verification. SSD~\cite{spector2023accelerating} restructures drafts into a tree and conducts batch decoding. SpecTr~\cite{sun2023spectr} seeks an optimal tradeoff between more draft candidates and the associated cost.

Our method belongs to speculative decoding that operates without external draft models. The recent study, Medusa~\cite{medusa}, shares similarities with BiTA in generating future tokens without altering original model parameters. However, a notable distinction lies in structure: BiTA employs soft embeddings, whereas Medusa utilizes multiple heads. Another recent study closely aligned with BiTA is PaSS~\cite{monea2023pass}, using ``look-ahead'' embeddings (referred as ``mask tokens'' in BiTA) for future predictions, while BiTA incorporates additional prompt tokens, which prove beneficial in experiments. Moreover, both works require calling the model a second time to validate draft candidates, while BiTA seamlessly conducts speculative generation and verification.

\subsection{Prompt Tuning}
As a widely adopted parameter-efficient tuning (PET) technique, Prompt Tuning~\cite{lester2021power}, along with various subsequent methods~\cite{li2021prefix,liu2023gpt}, optimizes pretrained transformers by updating a minimal set of prompt tokens, enhancing model customization for specific tasks, domains, or requirements. In this study, we leverage benefits of prompt tuning, introducing deep soft prompting from prefix tuning~\cite{li2021prefix} to effectively adapt AR language models for SAR decoding without modifying the original model parameters.

\begin{figure}[t]
\centering
\includegraphics[width=3.2in]{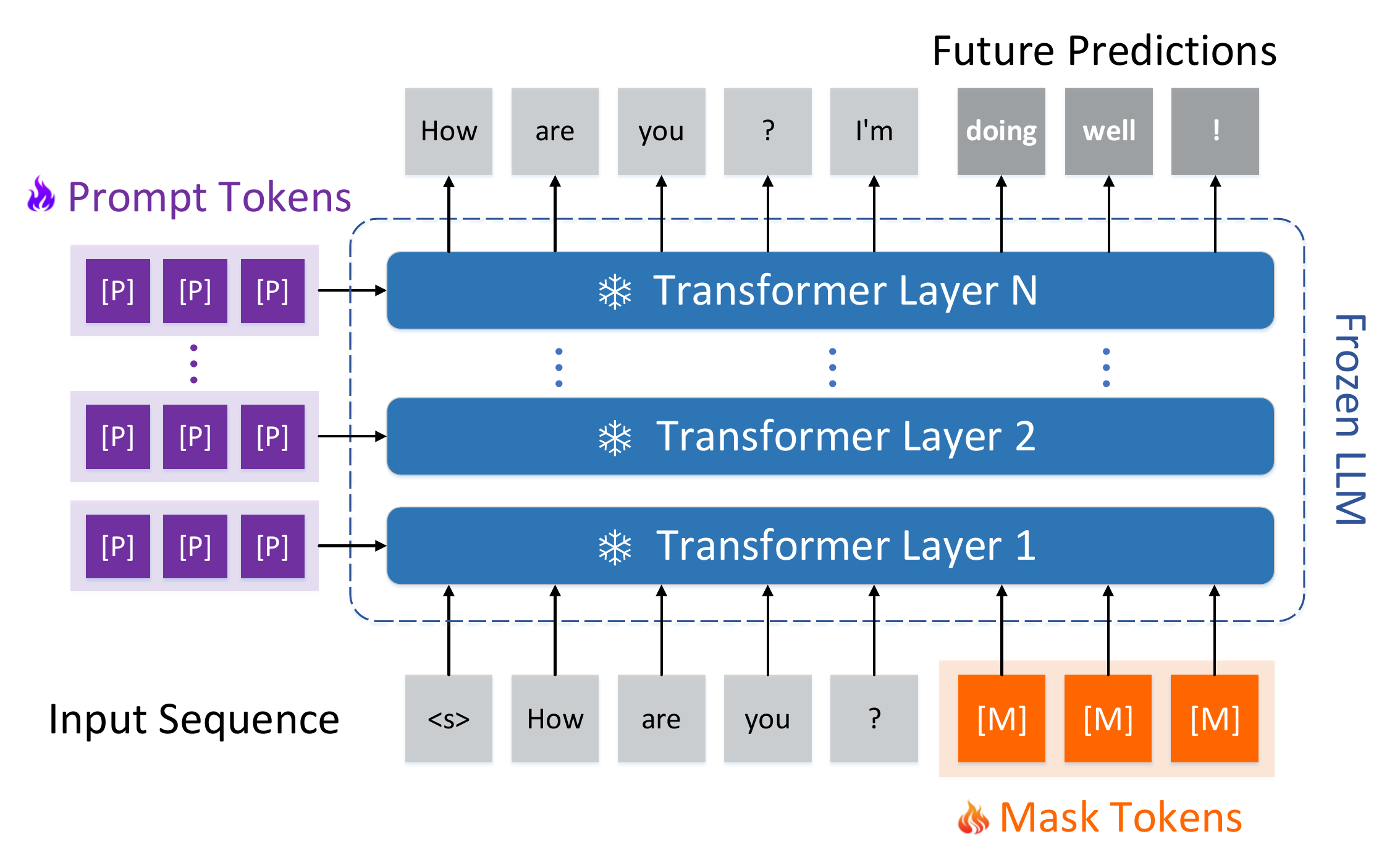}
\caption{A diagram of bi-directional tuning, orange blocks [M] for trainable mask tokens, purple blocks [P] for trainable prompt tokens, and blue blocks for transformer layers in frozen LLM. The predicted SAR future tokens are generated with the joint influence of frozen LLM parameters, prompt tokens, and mask tokens. For illustration purposes, we set the count of prompt and mask tokens to be 3.}
\label{fig:BiT-train}
\end{figure}

\section{Method}
In this section, we introduce BiTA, an innovative method for lossless LLM acceleration. Incorporating the proposed bi-directional tuning, BiTA enables the seamless adaptation of a transformer-based AR model to acquire an SAR generation style through efficient tuning. Additionally, we develop a streamlined generation and verification strategy facilitated by an efficient tree-based attention mechanism for lossless inference. The intricate details of these two pivotal components are expounded upon in the subsequent subsections.

\subsection{Bi-directional Tuning}\label{sec:bit}

\begin{figure}[t]
\centering
\includegraphics[width=2.7in]{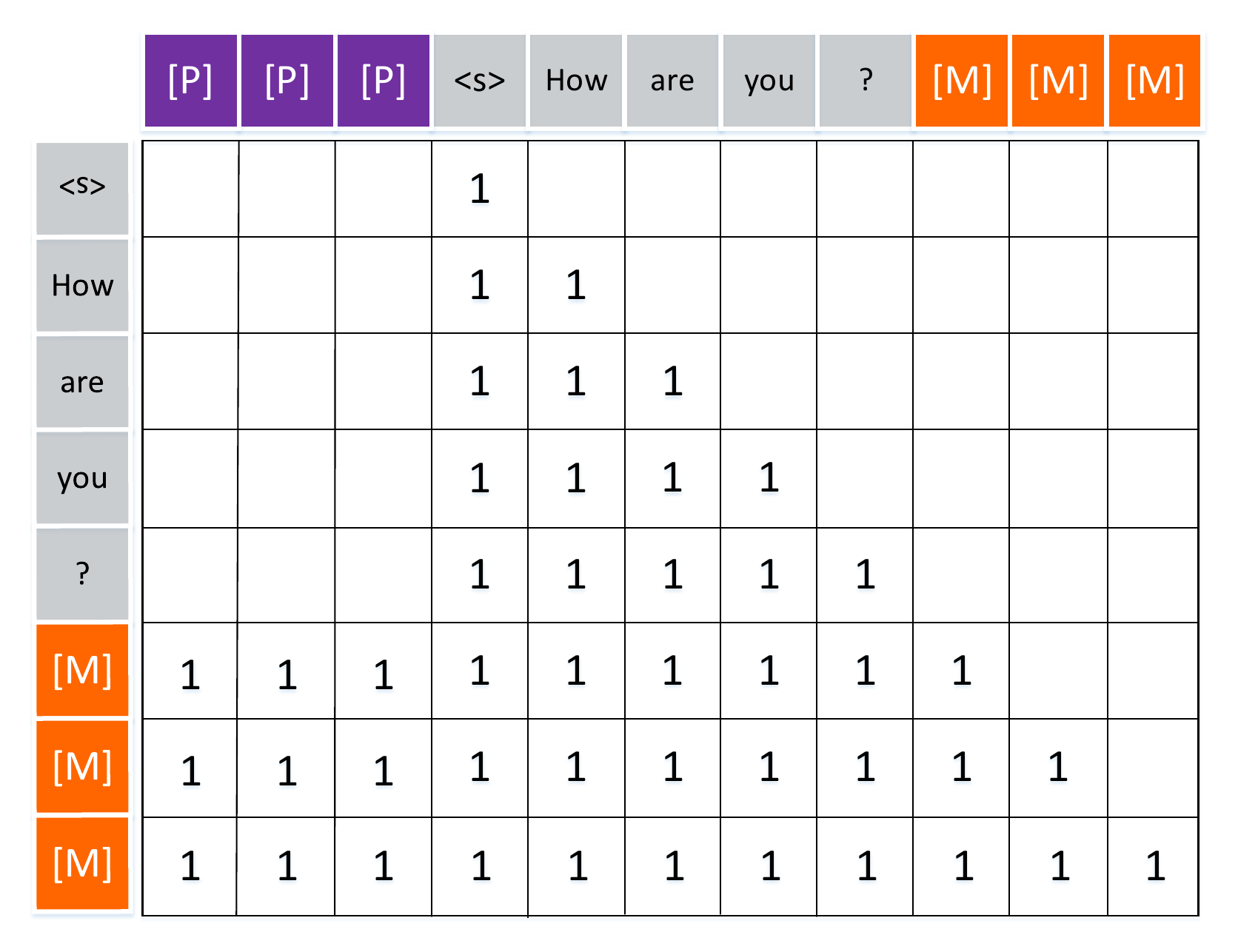}
\caption{An illustrative example of the attention mask employed in bi-directional tuning. The ``1'' indicates activation, while ``blank'' signifies suppression in attention mechanism. The shown example is derived from the sentence in Figure~\ref{fig:BiT-train}.}
\label{fig:BiT-attention}
\end{figure}


\begin{figure}[t]
\centering
\includegraphics[width=3.375in]{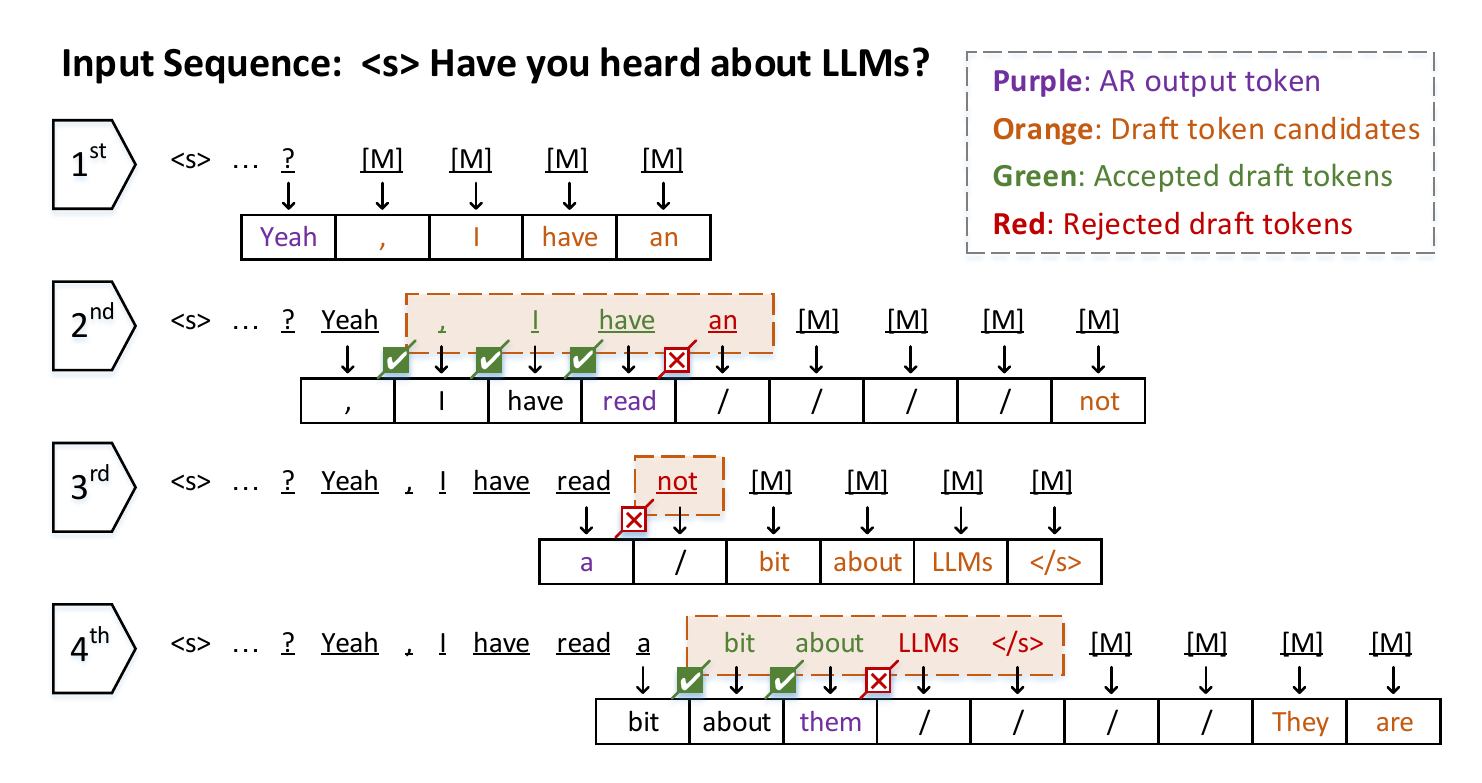}
\caption{A simple example illustrates the straightforward streamlined generation and verification. Input query, namely ``$<$s$>$ Have you heard about LLMs?'', serves as initial input token sequence $X^{0}$. The draft token candidates $\hat{C}^{i}$ are enclosed in orange dashed boxes. A successful acceptance is marked by a ``check''; otherwise, it is marked by a ``cross''. If $c$ draft tokens are accepted, the first $c$ mask tokens would be discarded as they are no longer necessary. If a draft candidate is rejected, its prediction, along with its subsequent tokens, is discarded (denoted as ``$/$''). In these four forward passes, the model produces 1, 4, 1, and 3 output tokens, respectively.}
\label{fig:naive}
\end{figure}

Thanks to the transformer architectures of LLMs, we leverage multiple learnable placeholder tokens known as mask tokens, empowering language models to generate subsequent consecutive future tokens beyond the last input token. During training and inference, the mask tokens are tasked with producing the probability of their next token in corresponding positions.

To maintain consistency with AR outputs, we choose to keep the original model parameters unchanged. Drawing inspiration from the idea of prompt tuning~\cite{li2021prefix}, we utilize prefix soft prompt tokens in the frozen LLMs, collaborating with the newly added mask tokens to achieve SAR generation (refer to Figure~\ref{fig:BiT-train} for a schematic diagram). In stark contrast, our design employs a distinct attention mechanism, specifically for the mask tokens, separate from the conventional input tokens. As illustrated in Figure~\ref{fig:BiT-attention}, the distinct attention allows prefix prompting to influence only the mask tokens rather than all tokens to their right, thereby avoiding alterations to the outputs of the original input tokens. To some extent, the prompt tokens serve as a specialized form of soft prompting, activating AR models to acquire the capability of SAR generation. 

During the training procedure, we perform bi-directional tuning in self-generated SFT-like instruction data, incorporating an SAR loss function. In the following, we describe how to produce the SAR training data and then introduce the used loss. Given a preprepared question token sequence $X$, we use the LLM intended for acceleration to generate answer token sequence $Y=\{y_0,~y_1,~...~,~y_N\}$ by greedy sampling. Thus, a question and the corresponding self-generated answer constitute a training sample. In a detailed explanation of training, considering the number of mask tokens as M, we randomly sample an index $k$ from the range of (0,~N$-$M). Subsequently, we form the question token sequence $X$ by appending it with $Y_{inst}^{k}=\{y_i,~i \leq k\}$ as a SAR task instruction. While the subsequent $Y_{gt}^{k,M}=\{y_i,~k+1 \leq i \leq k+M\}$ serves as its ground truth. Following the data production process above, we effortlessly manufacture a substantial amount of training data using preprepared question sequences from multiple SFT training datasets. Based on the SAR training data, the formulation of the SAR loss is as follows:
\begin{align}
\mathcal{L}_{SAR} &= - \sum_{i=k+1}^{k+M}logP(y_i|X, Y_{inst}^{k}; \theta, \phi_p, \phi_m),
\end{align}
where $y_i$ is the label of the mask tokens, drawn from $Y_{gt}^{k,M}$, $\theta$ denotes the frozen model parameters, and $\phi_p$, $\phi_m$ denotes the embeddings of prompt and mask tokens, respectively.

It is crucial to emphasize the use of self-generated training data, ensuring that the distribution of SAR output tokens closely aligns with that of the original AR outputs of the accelerated LLM. This simple yet effective strategy promotes accurate predictions, enhancing the model's ability to ``guessing'' future tokens precisely.

\subsection{Streamlined Generation \& Verification}\label{sec:streamlined}
BiTA simplifies SAR generation and verification in a single forward pass within a universal model. To illustrate, we first present a straightforward example of the streamlined generation and verification process. We then enhance it with more efficient decoding incorporating a tree-based attention mechanism inspired by literature~\cite{miao2023specinfer}.

An illustrative example of straightforward decoding is depicted in Figure~\ref{fig:naive}, and we describe the process as follows. Given an input token sequence $X^{0}=\{x_0,~x_1,~...,~x_{N-1}\}$, in the first forward pass, the model simultaneously outputs the prediction of the next token $\hat{y}_N$ and M future predictions $\hat{C}^{1} = \{\hat{y}_{N+1},~\hat{y}_{N+2},~...,~\hat{y}_{N+M}\}$. As a reliable AR output based on the context of $X^{0}$, $\hat{y}_N$ is indubitably accepted, while $\hat{C}^{1}$ serves as draft token candidates intended for verification in the next forward pass. Then, in the $i_{th}~(i > 1)$ forward pass, the input sequence comprises query tokens, previously generated tokens, draft token candidates $\hat{C}^{i-1}$ (enclosed in orange dashed boxes in Figure~\ref{fig:naive}), and mask tokens. The draft token candidates are sequentially verified based on the AR prediction of the last token. The accepted draft candidates are added to the previously generated token sequence. If a draft candidate is rejected, its prediction, along with its follow-ups, is discarded, and the prediction of the last accepted candidate serves as the AR output token which is accepted. For draft generation, each mask token outputs the prediction of its next token in corresponding positions. Note that they do \textbf{not} ``attend'' to the current draft token candidates, generating future predictions in the context of only both query tokens and previously generated tokens. The predicted future tokens are used in the the next forward pass for verification.

\begin{figure}[t]
\centering
\includegraphics[width=3.3in]{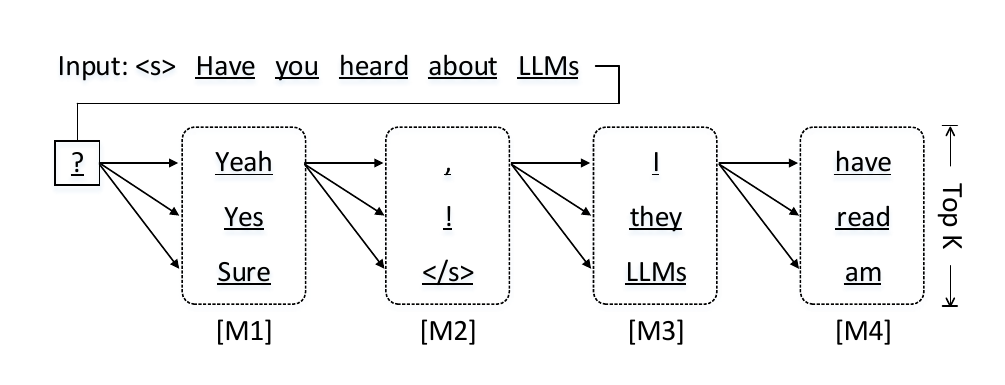}
\vspace{-7.5pt}
\caption{The efficient draft candidate token tree. The configuration includes 4 mask tokens, and for the prediction of each mask token, the top 3 draft candidates are selected. As shown, only the top-1 scoring word has subsequent words for verification.}
\label{fig:tree}
\end{figure}

The straightforward streamlined generation and verification method faces two tricky challenges. On one hand, the correctness of draft token candidates is not always assured. As stated, the draft token candidates are predicted based on the context of the output sequence until the last forward pass, rather than incorporating the currently accepted draft tokens. On the other hand, the number of draft token candidates may be very few, as some draft token candidates are no longer effective due to being accepted in the last forward pass, or even zero if all candidates are accepted in the last forward pass. These two inadequacies become significant obstacles for decoding efficiency. Moreover, while the draft token candidates are currently drawn from the top-1 scoring word of future predictions, exploring additional likely words for each prediction could potentially increase the probability of acceptance, leading to a higher inference speedup~\cite{xia2023speculative}.

To overcome these identified inadequacies, we present an efficient tree-based decoding approach using an elaborate attention mechanism to ensure an adequate number and correctness of draft token candidates. For each mask token, we select the top-k predicted words as draft candidates in corresponding positions. All groups of top-k draft candidates can be organized into a token tree, where each leaf node indicates a possible generated sequence for verification. By exploring the distribution of the output future tokens, we observe that the top-1 scoring word plays the most important role in subsequent predictions: the most likely candidate has a higher chance of forming an effective sequence with the subsequent words. Therefore, instead of building a fully tree to explore all possible candidate sequences, we construct an efficient token tree that is sufficiently representative, concentrating on those sequences whose non-leaf nodes in the token tree are the top-1 scoring candidate, as illustrated in Figure~\ref{fig:tree}. 

\begin{figure*}[t]
\centering
\includegraphics[width=6.9in]{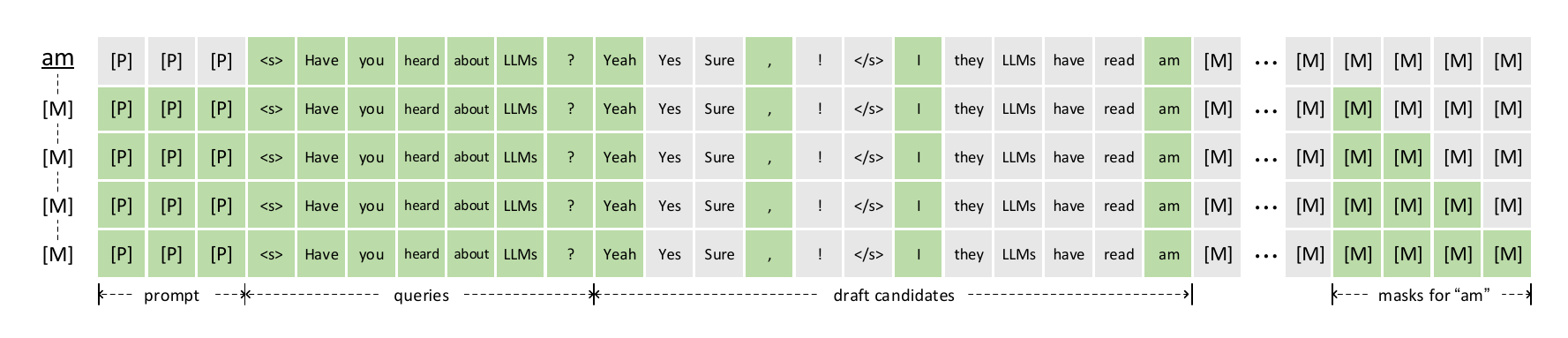}
\vspace{-7.5pt}
\caption{Partial attention mask in efficient tree-based decoding. Activation is indicated in green, while suppression is shown in gray. The presented attention mask illustrates the attention of the candidate word ``am'' and the corresponding attached mask tokens. The word ``am'' is derived from the last layer of the token tree in Figure~\ref{fig:tree}. Based on the connection of nodes, the word ``am'', along with the attached mask tokens, ``attends'' to the candidate words ``I'', ``,'', and ``Yeah'' as depicted.}
\label{fig:tree-attention}
\end{figure*}

Moreover, we attach a group of mask tokens to each node in the efficient token tree. This ensures that regardless of the number of currently accepted draft candidates, we can always select the same number of mask tokens that follows the last accepted candidate token. The selected mask tokens are utilized to organize the draft candidate token tree for the next step. By having the selected mask tokens predict future tokens based on the context of the accepted candidate tokens, there is a potential improvement in accuracy. This improvement is expected to result in higher efficiency for decoding, as demonstrated in Section~\ref{sec:super}.

To obtain a sequential input, we flatten the draft candidate token tree, along with the attached mask tokens, into a token sequence. This sequence is then appended to the current result token sequence which consists of query tokens and previously generated tokens, to construct the model input. During inference, the draft candidate tokens and the attached mask tokens ``attend'' to both their ``ancestors'' in the tree and the current result tokens as well, while the mask tokens ``attend'' to prompt tokens additionally. Thanks to the tree-based attention, AR and SAR generation, as well as draft verification, are conducted in a single forward pass simultaneously. An incomplete yet sufficiently expressive illustration of the tree-based attention is provided in Figure~\ref{fig:tree-attention} (the shown example is derived from the token tree depicted in Figure~\ref{fig:tree}).

In this subsection, we delineate the streamlined decoding method step by step. Beginning with a straightforward simplified toy example, we introduce the concept of streamlined generation and verification in parallel. In order to improve acceptance rate of drafts, we design an efficient top-k draft candidate token tree with candidate-wise mask token groups. Finally, we employ a tree-based attention mechanism to achieve streamlined generation and verification. Despite introducing additional computational complexity with more tokens used in inference, this strategy still achieves an impressive speedup in LLMs. Because the bottleneck of LLM's inference is usually the memory-bandwidth cost~\cite{shazeer2019fast} rather than computational cost. Additionally, we explore the size of the draft candidate token tree, achieving a favorable acceptance-complexity tradeoff (detailed in Section~\ref{sec:ab}).

\section{Experiments}
\subsection{Experimental setup}
\subsubsection{Datasets}
As described in Section~\ref{sec:bit}, we utilize self-generated SFT-like training data, which comprises preprepared questions and the answers generated by the LLM for acceleration. For diversity, the preprepared questions are sourced from five SFT datasets: LIMA~\cite{zhou2023lima}, Alpaca-GPT4~\cite{peng2023instruction}, CodeAlpaca~\cite{codealpaca}, OpenPlatypus~\cite{platypus2023}, and CIP~\cite{cip2023}. Note that we use a $50k$ sample set from the CIP training set during training, resulting in approximately $150k$ training samples in total.

For evaluation, we employ four datasets: XSum~\cite{narayan2018don}, MT-Bench~\cite{zheng2023judging}, the CIP test set, and HumanEval-X~\cite{zheng2023codegeex}. This evaluation covers the acceleration capability across summarization, open-ended questions, conversation, and code, respectively.

\subsubsection{Implementation Details}
Experiments are conducted on LLMs of various sizes, encompassing the LLaMA-2 chat model series (7B, 13B, 70B)~\cite{touvron2023llama2}, Vicuna series (7B, 13B, 33B)~\cite{vicuna2023}, and Falcon-40B-Chat~\cite{almazrouei2023falcon}, ranging from 7B to 70B parameters. For optimization, a cosine learning rate schedule is employed with an initial value of 3e-2 and a batch size of 128 for 4 epochs. The learnable embeddings are initialized with a normal distribution having a mean of zero and a standard deviation of 0.02. Unless specified otherwise, the count of prompt tokens is established at 16, with mask tokens set to 3 for models having a parameter size not exceeding 13B and adjusted to 4 otherwise. The training is conducted on a cluster of four servers, each equipped with eight NVIDIA A800 (80GB) GPUs. For a base model with a scale of 7B parameters, the training process takes approximately 7 hours. In addition, for our efficient tree-based decoding, the top 5 predictions in each mask token are selected as draft token candidates.

\subsubsection{Evaluation Settings}
We conduct the evaluation of the proposed BiTA and comparative methods on NVIDIA A800 (80GB) GPUs. For models with a parameter size no greater than 13B, a single A800 GPU is used; otherwise, 8 A800 GPUs are used. The evaluated models performed inference with a batch size of 1 to ensure accurate assessment of response time. To facilitate comparison, we use ``greedy speedup'' as the metric~\cite{medusa}, defined as the ratio of the evaluated model's speed using greedy sampling to the AR baseline, with speed measured in generated tokens per second. The AR baseline is implemented using the AR generation code in the Huggingface Transformers library~\cite{wolf-etal-2020-transformers}.

\subsection{Main Results}
\subsubsection{Speedup Across Diverse LLMs and Tasks}
Table~\ref{tab:main1} presents the speedup of the proposed BiTA across four datasets: XSum, MT-Bench, CIP, and HumanEval-X. When BiTA is applied, the expedited LLMs exhibit a speedup ranging from $2.1\times$ to $3.3\times$ across various generation tasks, encompassing summarization, open-ended questions, conversation, and code. Notably, larger LLMs tend to exhibit more substantial speedup, possibly attributed to the intrinsic richer context encoded by the embeddings for each token, facilitating improved future predictions. Another hypothesis is that our prompting-based methods benefit from larger models, aligning with similar observations in NLP tasks discussed in other works on parameter-efficient tuning~\cite{lester2021power,liu2023gpt}. Additionally, it is noteworthy that BiTA attains particularly impressive speedup ($2.7$$\sim$$3.3\times$) in code generation compared to other tasks. Upon examining the test samples, we hypothesize that the structured and logical content in code generation tasks may play a significant role in enhancing future predictions. In addition to the remarkable speedup performance, we observe that for a 7B-scale base model, the count of trainable parameters for prompt and mask embeddings is approximately $0.06\%$ of the total model parameters, while it is 10 times less for 70B-scale models.

\begin{table}[!tbp]
\centering
\small
\begin{tabular}{p{60pt}p{30pt}<{\centering}p{30pt}<{\centering}p{30pt}<{\centering}p{30pt}<{\centering}}
\hline
Model & \multicolumn{1}{c}{XSum} & \multicolumn{1}{c}{MT-B} & \multicolumn{1}{c}{CIP} & \multicolumn{1}{c}{HE-X} \\
\hline
LLaMA-2-7B  & 2.19 & 2.38 & 2.29 & 2.73 \\
LLaMA-2-13B & 2.29 & 2.41 & 2.39 & 2.88 \\
Vicuna-33B  & 2.20 & 2.47 & 2.10 & 3.00 \\
Falcon-40B & 2.28 & \textbf{2.75} & 2.32 & 3.07 \\
LLaMA-2-70B & \textbf{2.55} & 2.72 & \textbf{2.58} & \textbf{3.31} \\
\hline
\end{tabular}
\caption{The speedup of BiTA in various base models on XSum, MT-Bench, CIP, and HumanEval-X under greedy sampling setting. For space conservation, MT-Bench is abbreviated as MT-B, and HumanEval-X is shortened to HE-X. The involved models are chat versions, which are not further explained in the rest of the paper.}
\label{tab:main1}
\end{table}

\subsubsection{Comparison with Speculative Decoding}
We compare the speedup performance of our method with several state-of-the-art speculative decoding based methods in Table~\ref{tab:main2}. The evaluation is conducted on MT-Bench, utilizing both LLaMA-2-7B and LLaMA-2-70B. For a fair comparison, we implement the four comparative approaches using publicly available source codes in the same hardware environment environment as our BiTA. As these four comparative approaches do not require special training or finetuning, we can easily re-implement them in our experimental environment. However, for HF AssistGen~\cite{gante2023assisted} and SpecDec~\cite{leviathan2023fast}, we encountered challenges in acquiring a compact draft model with a similar output token distribution to the target model, thus failing to meet the acceleration requirement for LLaMA-2-7B, and we cannot provide the corresponding results. To expedite LLaMA-2-70B, we use LLaMA-2-7B as the draft model for them. As shown in Table~\ref{tab:main2}, BiTA achieves a speedup of up to $2.7\times$, significantly outperforming the comparative methods.

\begin{figure}[t]
\centering
\includegraphics[width=3.0in]{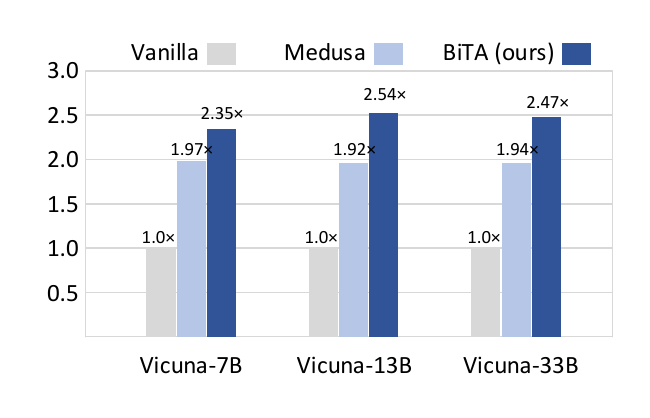}
\vspace{-10pt}
\caption{A comparison on MT-Bench between Medusa and BiTA using greedy sampling. The reported Medusa speedup is considered.}
\label{fig:speedup}
\end{figure}

In addition to comparing with the four speculative decoding methods mentioned above, we also assess BiTA against a recent study, Medusa~\cite{medusa}, because of its similar motivation to our approach for SAR generation and verification. We apply BiTA to Vicuna-7B, 13B, and 33B on MT-Bench to compare the speedup performance reported in Medusa. Figure~\ref{fig:speedup} demonstrates that our method outperforms Medusa with an improvement ranging from $19$$\sim$$32\%$ across various base models (though it is important to note that the gap between the two methods may not be entirely reliable due to potential differences in experimental settings). We attribute the superiority of our method to its powerful bi-directional tuning, where mask tokens can capture a richer feature context during the forward pass. Furthermore, the simultaneous generation and verification strategy contribute to the acceleration as well.

\begin{table}[!tbp]
\centering
\small
\begin{tabular}{p{142pt}p{30pt}<{\centering}p{30pt}<{\centering}}
\hline
Acceleration methods in LLaMA-2 & 7B & 70B \\
\hline
HF AssistGen~\cite{gante2023assisted} & - & 1.53  \\
SpecDec~\cite{leviathan2023fast} & - & 1.79 \\
Self-SpecDec~\cite{zhang2023draft} & 1.11 & 1.15 \\
Lookahead~\cite{fu2023lookahead} & 1.70 & 1.22 \\
\hline
BiTA (ours) & \textbf{2.38} & \textbf{2.72} \\
\hline
\end{tabular}
\caption{Comparison of speedup between BiTA and speculative decoding based acceleration approaches on MT-Bench.}
\label{tab:main2}
\end{table}

\subsection{Ablation Study}\label{sec:ab}
\subsubsection{Impact of Prompting Design}
To validate effectiveness of the bi-directional tuning design, we explore the impact of various prompting designs, including no prompting, shallow prompt tuning, and deep prompt tuning. The comparative experiments are conducted using LLaMA-2-7B for conversation (CIP) and code (HumanEval-X) generation tasks. As indicated in Table~\ref{tab:ab1}, we observe a progressive speedup gain. When utilizing only mask tokens for future prediction, a significantly lower speedup ($1.94\times$ on CIP and $2.12\times$ on HumanEval-X) is observed. However, when adding prompt tokens as learnable soft embeddings in the input sequence~\cite{lester2021power}, referred to as shallow prompt tuning, the speedup performance improves slightly ($+0.07$ on CIP and $+0.19$ on HumanEval-X), but it falls short of the level achieved with bi-directional tuning ($2.29\times$ on CIP and $2.73\times$ on HumanEval-X), where soft prompt tuning is applied in every transformer block.

\subsubsection{Speedup \emph{vs.} Prompt Tokens}
\begin{figure}[t]
\centering
\includegraphics[width=3.3in]{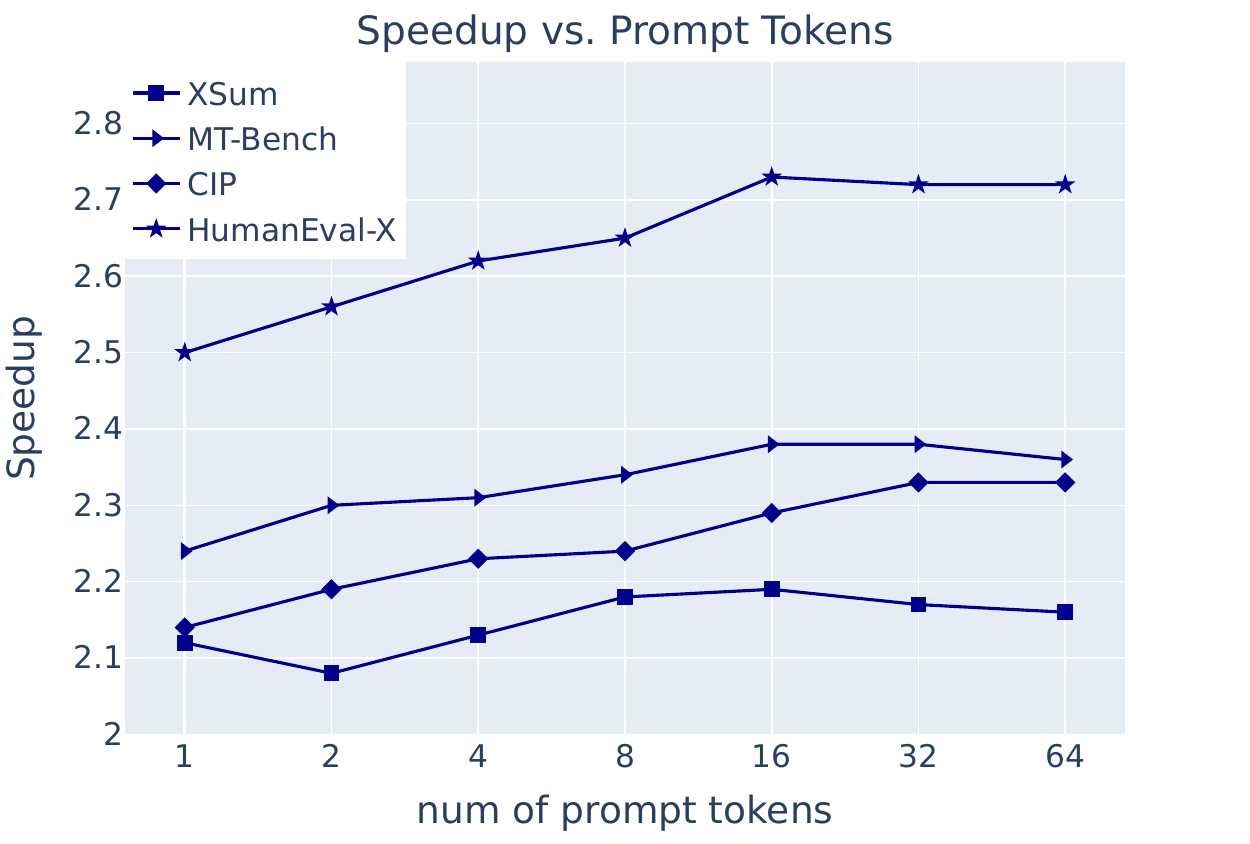}
\vspace{0pt}
\caption{Speedup \emph{vs.} the number of prompt tokens. Various dataset speedups are represented by distinct markers, with LLaMA-2-7B as the base model.}
\label{fig:prompt}
\end{figure}

Obviously, using more prompt tokens involves more trainable model parameters, resulting in a slight increase in computational complexity during inference. We explore the influence of the number of prompt tokens to understand its influence on speed. Our investigation focuses on LLaMA-2-7B across four datasets: XSum, MT-Bench, CIP, and HumanEval-X. In Figure~\ref{fig:prompt}, we observe that the speedup approximately increases with the number of used prompt tokens, consistently across all four datasets, and reaches saturation at 16. Interestingly, the speedup with 8$\sim$64 prompt tokens is not significantly different. This highlights that through bi-directional tuning, once a specific threshold number of learnable deep prompt embeddings are incorporated, AR models can be effectively adapted into an SAR style.

\begin{figure}[t]
\centering
\includegraphics[width=3.3in]{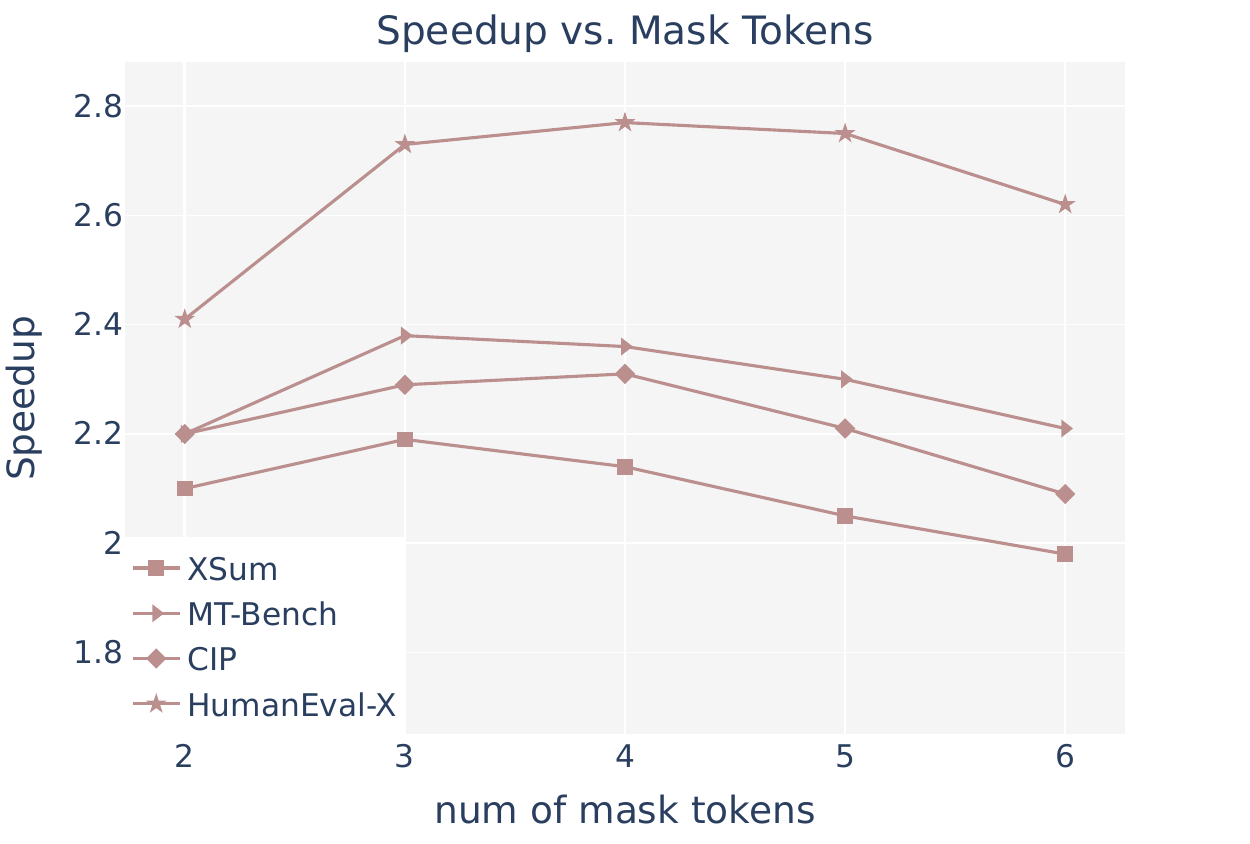}
\vspace{0pt}
\caption{Speedup \emph{vs.} the number of mask tokens. Various dataset speedups are represented by distinct markers, with LLaMA-2-7B as the base model.}
\label{fig:mask}
\end{figure}

\subsubsection{Speedup \emph{vs.} Mask Tokens}
Increasing the number of mask tokens provides more draft token candidates, enhancing the potential to generate additional output tokens for SAR decoding in a single forward step. However, a higher number of draft candidates introduces heavier computational overhead, slowing down model inference. An ablation study explores the impact of varying the number of masked tokens, as presented in Figure~\ref{fig:mask}. The experiments are conducted using LLaMA-2-7B. The observation of speedup across four datasets indicates that choosing 3 or 4 mask tokens achieves a favorable trade-off between SAR decoding capacity and computational overhead. Thus, through exhaustive experimental search, we set the number of mask tokens to 3 for models with no more than 13 billion parameters and to 4 for larger models, aiming for the highest speedup across diverse generation tasks.

\subsubsection{Superiority of Efficient Tree-based Decoding}\label{sec:super}
As explained in Section~\ref{sec:streamlined}, we develop the efficient tree-based decoding starting from a straightforward simplified decoding method. To demonstrate the efficiency of the proposed method, we compare the efficient tree-based decoding with the straightforward decoding and the conventional fully tree-based decoding. The speedup achieved with different decoding methods based on LLaMA-2-7B is presented in Table~\ref{tab:ab4}. For fully tree-based decoding, we conducted a comprehensive exploration of the number $k$ for the top-k predictions in each mask token, as draft candidates. The speedup results indicate that decoding with an efficient token tree consistently outperforms decoding with a fully token tree in all cases of $k$ on both XSum and MT-Bench, surpassing the straightforward decoding with an improvement exceeding $30\%$.

\begin{table}[!tbp]
\centering
\small
\begin{tabular}{p{98pt}p{50pt}p{55pt}}
\hline
Different Prompting & CIP & HumanEval-X \\
\hline
mask tokens only & 1.94 & ~2.12  \\
~~ + shallow prompt tokens & 2.01 (+0.07) & ~2.31 (+0.19) \\
bi-directional tuning & 2.29 (+0.35) & ~2.73 (+0.61) \\
\hline
\end{tabular}
\caption{Speedup achieved with different prompting designs on CIP and HumanEval-X, using LLaMA-2-7B as the base model.}
\label{tab:ab1}
\end{table}

\begin{table}[!tbp]
\centering
\small
\begin{tabular}{p{133pt}p{25pt}<{\centering}p{35pt}<{\centering}}
\hline
Decoding Method & \multicolumn{1}{c}{XSum} & \multicolumn{1}{c}{MT-Bench} \\
\hline\hline
Straightforward Decoding   & 1.63 & 1.82 \\
\hline
Fully Tree-based Decoding ($k=1$) & 1.81 & 2.04 \\
Fully Tree-based Decoding ($k=2$)  & 2.02 & 2.23 \\
Fully Tree-based Decoding ($k=3$) & 1.97 & 2.05 \\
Fully Tree-based Decoding ($k=4$) & 1.69 & 1.64 \\
\hline
Efficient Tree-based Decoding (ours) & 2.19 & 2.38 \\
\hline
\end{tabular}
\caption{Speedup achieved with various decoding method on XSum and MT-Bench, using LLaMA-2-7B as the base model. The number $k$ denotes top $k$ predicted words used as draft candidates for each mask token.}
\label{tab:ab4}
\end{table}

\section{Conclusion}
We present a novel method named BiTA for achieving lossless acceleration in LLMs. To reduce transformer calls in autoregressive LLMs during inference, BiTA seamlessly adapts existing AR models for an SAR generation style through the proposed bi-directional tuning, utilizing very limited trainable parameters. Based on the tree-based efficient decoding strategy, the model conducts streamlined generation and verification simultaneously. These two features of BiTA jointly contribute to expediting LLMs without altering the original outputs. Extensive experimental results demonstrate a remarkable speedup ranging from $2.1\times$ to $3.3\times$ across LLMs of various sizes and diverse generation tasks. Furthermore, due to its flexible prompting design, BiTA serve as a plug-and-play technique applicable to any publicly available LLMs for acceleration, which is of great significance in resource-limited scenarios and real-time applications.

\clearpage
\appendix

\bibliographystyle{named}
\bibliography{ijcai24}

\onecolumn
\section{Details on Self-Generated Training Data}
We elaborate on the process of generating the self-generated SFT-like training data used in BiTA. The SFT training datasets utilized are enumerated in Table~\ref{tab:dataset}. It is important to note that only the questions, also referred to as ``queries'' or ``instructions,'' from these datasets are utilized. These questions are standardized through predefined prompt templates and are then input into the LLM intended for acceleration. The answers generated through greedy sampling serve as the ground truth, employed in the training procedure. For efficiency purposes, the LLM is deployed in the Text Generation Inference (TGI) framework. 

The employed prompt templates are outlined in Table~\ref{tab:template}. We utilize two types of prompt templates for both LLaMA-2 based models and Vicuna based models, accommodating different use cases. The distinction between the two templates lies in whether to incorporate system information. Consequently, a single question is input into both templates, leading to the generation of two distinct answers. In the case of Falcon, a sole prompt template suffices, as recommended by Falcon's Hugging Face community for simplicity. Following the generation of training data, the questions and answers, along with their corresponding prompt templates, constitute training samples. Due to the quantity of questions utilized, there are approximately 300k training samples per training epoch for LLaMA-2 and Vicuna, whereas only 150k for Falcon. To ensure consistency in model training, we opt for a straightforward approach of doubling the training epochs (\emph{i.e.} $4 \rightarrow 8$) for Falcon.

\begin{table*}[htbp]
\small
\centering
\begin{tabular}{p{60pt}p{50pt}<{\centering}p{350pt}}
\hline
~Dataset & \# Sample & ~~~~~~~~~~~~~~~~~~~~~~~~~~~~~~~~~~~~~~~~~~~~~~~~~~~~Description \\
\hline
~LIMA & 1,029 & carefully curated high-quality data \\
~Alpaca-GPT4 & 52,002 & instruction-following data generated by GPT-4 \\
~CodeAlpaca & 20,022 & instruction-following code data \\
~OpenPlatypus & 24,926 & a curated dataset derived from 11 open-source datasets, with a focus on STEM and logic\\
~CIP (train set)& 257,999 & a conversational dataset spanning various domains\\
\hline
\end{tabular}
\caption{Statistics of the SFT datasets associated with the training data generation. Please note that we exclusively utilize a 50k sample set from CIP, resulting in a total of approximately 150k questions being employed.}
\label{tab:dataset}
\end{table*}

\begin{table*}[htbp]
\small
\centering
\begin{tabular}{p{50pt}p{25pt}<{\centering}p{385pt}}
\hline
Model & Type & ~~~~~~~~~~~~~~~~~~~~~~~~~~~~~~~~~~~~~~~~~~~~~~~~~~~~~~~~~~~~~~~~~~~~Template \\ \hline
\multirow{7}{*}{LLaMA-2} & \multirow{6}{*}{full} & $<$s$>$[INST] $<<$SYS$>>$$\backslash$nYou are a helpful, respectful and honest assistant. Always answer as helpfully as possible, while being safe.~~Your answers should not include any harmful, unethical, racist, sexist, toxic, dangerous, or illegal content. Please ensure that your responses are socially unbiased and positive in nature.$\backslash$n$\backslash$nIf a question does not make any sense, or is not factually coherent, explain why instead of answering something not correct. If you don't know the answer to a question, please don't share false information.$\backslash$n$<<$/SYS$>>$$\backslash$n$\backslash$n\{\textbf{Question}\} [/INST]~\\ \cline{2-3} 
& short & $<$s$>$[INST] \{\textbf{Question}\} [/INST]~ \\ \hline
\multirow{3}{*}{Vicuna} & \multirow{2}{*}{full} & $<$s$>$A chat between a curious user and an artificial intelligence assistant. The assistant gives helpful, detailed, and polite answers to the user's questions. USER: \{\textbf{Question}\} ASSISTANT: \\ \cline{2-3} 
& short & $<$s$>$USER: \{\textbf{Question}\} ASSISTANT: \\ \hline
Falcon & full & User: \{\textbf{Question}\}$\backslash$nAssistant: \\ \hline
\end{tabular}
\caption{The utilized prompt templates for LLaMA-2 chat model series (7B, 13B, 70B), Vicuna series (7B, 13B, 33B), and Falcon-40B-Chat. \{\textbf{Question}\} represents the location where the question is placed.}
\label{tab:template}
\end{table*}

\section{Additional Details on Efficient Tree-Based Decoding}
\begin{figure*}[htbp]
\centering
\includegraphics[width=6.9in]{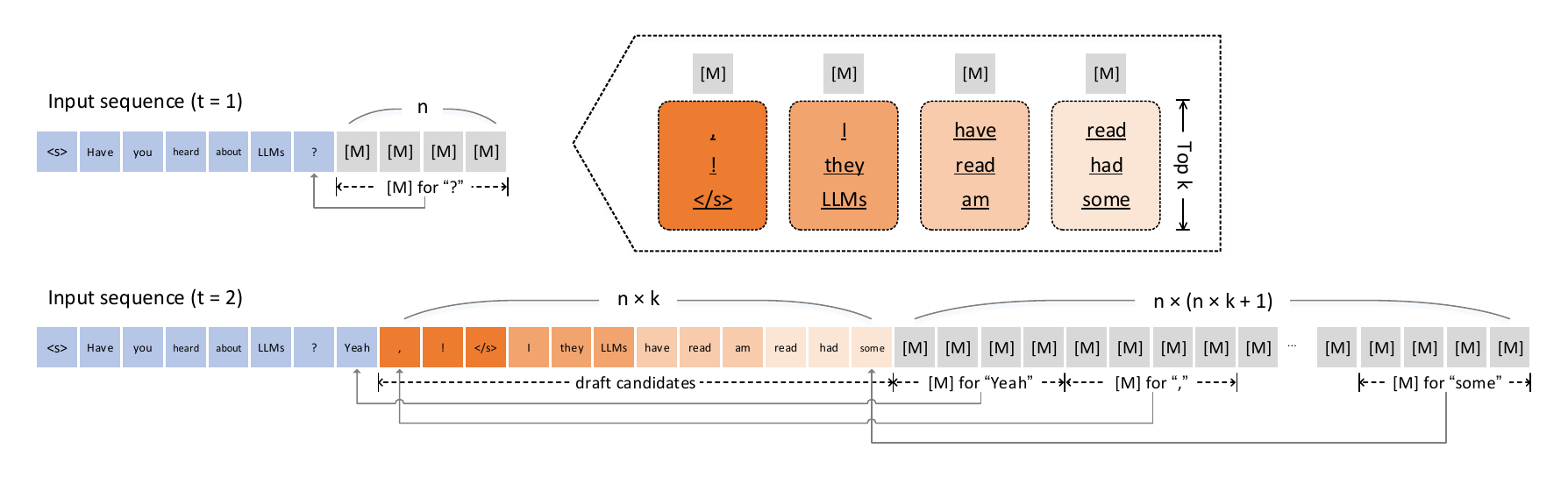}
\vspace{0pt}
\caption{A concrete illustration depicting input token sequences during the initial two forward passes in the efficient tree-based decoding. There are 4 mask tokens in each mask token group (denoted as $n=4$), and the top 3 predictions are chosen as draft candidates (denoted as $k=3$). The gray arrows signify that the mask token group is linked to the target token through attention mechanism.}
\label{fig:sample}
\end{figure*}

To provide a more detailed illustration of the efficient tree-based decoding process, we showcase the input token sequences during the first two forward passes in this decoding approach. As depicted in Figure~\ref{fig:sample}, during the first forward pass, we utilize the original input token sequence, commonly referred to as ``queries,'' along with the embeddings of $n$ mask tokens as input. Thus, each mask token predicts the logits of the next token in the corresponding positions. The top $k$ predictions for each mask token are selected as draft candidates, which are then constructed into an efficient token tree, as detailed in Section 3.2 of the main text. In the subsequent forward pass, the draft candidate token tree is flattened along its depth and appended to the current input token sequence which comprises the original input tokens and the newly generated tokens. For each draft candidate and the last token of the current input sequence, a mask token group is prepared, consisting of $n$ mask tokens for predicting their future tokens. This ensures that if a token is the last accepted token in this step, its future tokens will generate the draft candidates for the next forward step.

We employ the tree-based attention mechanism outlined in Section~3.2 of the main text to maintain the relevant context within the flattened input token sequence. In addition, Algorithm~\ref{alg:infer}, presented alongside the illustration of input token sequences, details the efficient tree-based decoding process as described in Section~3.2. 

\section{Additional Experimental Results}
\subsection{Speedup on MT-Bench for Various LLM Scales}
To provide a more detailed demonstration of the acceleration capability of our proposed method BiTA in various text generation tasks, we present the speedup based on different scales of LLMs on MT-Bench. Refer to Table~\ref{tab:mt} for details.

\begin{table*}[htbp]
\centering
\small
\begin{tabular}{p{60pt}|p{33pt}<{\centering}|p{37pt}<{\centering}|p{41pt}<{\centering}|p{33pt}<{\centering}|p{37pt}<{\centering}|p{37pt}<{\centering}|p{33pt}<{\centering}|p{33pt}<{\centering}|p{33pt}<{\centering}}
\hline
Model & Code & Extraction & Humanities & Math & Reasoning & Roleplay & STEM & Writing & Overall \\
\hline
LLaMA-2-7B  & 2.68 & 2.94 & 2.21 & 2.69 & 2.33 & 2.11 & 2.37 & 2.26 & 2.38 \\
Vicuna-7B  & 2.57 & 3.00 & 2.28 & 2.46 & 2.23 & 1.90 & 2.25 & 2.23 & 2.35 \\
LLaMA-2-13B & 2.67 & 2.96 & 2.35 & 2.73 & 2.37 & 2.13 & 2.46 & 2.06 & 2.41 \\
Vicuna-13B  & 2.76 & 2.88 & 2.32 & 2.90 & 2.58 & 2.18 & 2.36 & 2.48 & 2.54 \\
Vicuna-33B  & 3.24 & 3.02 & 2.23 & 3.32 & 2.40 & 2.17 & 2.21 & 2.23 & 2.47 \\
Falcon-40B  & 2.93 & 2.74 & 2.67 & 3.43 & 2.95 & 2.78 & 2.31 & 2.52 & 2.75 \\
LLaMA-2-70B  & 3.15 & 3.26 & 2.51 & 3.27 & 2.60 & 2.49 & 2.70 & 2.39 & 2.72 \\
\hline
\end{tabular}
\caption{The specific speedup in each subtask on MT-Bench under the greedy sampling setting.}
\label{tab:mt}
\end{table*}

\subsection{Impact of Prompting Design}
Table~\ref{tab:design} provides comparative experiments on the impacts of prompting design on the XSum and MT-Bench datasets, complementing Table 3 in the main text. The speedup improvements are consistent with those presented in the main text.

\begin{table*}[htbp]
\small
\centering
\begin{tabular}{p{125pt}p{72pt}p{72pt}}
\hline
Different Prompting & XSum & MT-Bench \\
\hline
mask tokens only & ~1.95 & ~2.00  \\
~~ + shallow prompt tokens & ~2.03 (+0.08) & ~2.06 (+0.06) \\
bi-directional tuning & ~2.19 (+0.24) & ~2.38 (+0.38) \\
\hline
\end{tabular}
\caption{Speedup achieved with different prompting designs on XSum and MT-Bench, using LLaMA-2-7B as the base model.}
\label{tab:design}
\end{table*}

\subsection{Superiority of Efficient Tree-based Decoding}
Table~\ref{tab:decoding} reinforces the effectiveness of our efficient tree-based decoding on both CIP and HumanEval-X, consistent with our findings on speedup across various decoding methods in Table 4 of the main text.

\begin{table*}[htbp]
\small
\centering
\begin{tabular}{p{160pt}p{50pt}<{\centering}p{60pt}<{\centering}}
\hline
Decoding Method & \multicolumn{1}{c}{CIP} & \multicolumn{1}{c}{HumanEval-X} \\
\hline\hline
Straightforward Decoding   & 1.89 & 2.06 \\
\hline
Fully Tree-based Decoding ($k=1$) & 2.02 & 2.39 \\
Fully Tree-based Decoding ($k=2$)  & 2.14 & 2.58 \\
Fully Tree-based Decoding ($k=3$) & 1.93 & 2.29 \\
Fully Tree-based Decoding ($k=4$) & 1.50 & 1.79 \\
\hline
Efficient Tree-based Decoding (ours) & 2.29 & 2.73 \\
\hline
\end{tabular}
\caption{Speedup achieved with various decoding method on CIP and HumanEval-X, using LLaMA-2-7B as the base model. The number $k$ denotes top $k$ predicted words used as draft candidates for each mask token.}
\label{tab:decoding}
\end{table*}

\begin{algorithm}[tbp]
\caption{Streamlined Generation \& Verification in Parallel}
\label{alg:infer}
\begin{algorithmic}[1]
\Require An input token sequence $\mathcal{T}$, number of mask tokens $n$, number of prediction candidates for each mask token $k$, the BiTA-enhanced autoregressive large language model $\mathcal{F}$, embeddings of prompt tokens $\mathcal{P}$, embeddings of mask tokens $\mathcal{M}$
\renewcommand{\algorithmicrequire}{\textbf{Declare}}
\Require Candidate set $\mathcal{C}^{t}=\{c_{i,j}^{t}, i\in[1,~n], j\in[1,~k]\}$ where $c_{i,j}^{t}$ denotes the $j_{th}$ prediction of the $i_{th}$ mask token $m_i$ at step $t$
\Ensure A generated token sequence $\mathcal{O}$
\State $\mathcal{O}=\varnothing$, $\mathcal{C}^{0}=\varnothing$
\State $t=1$
\While{True}
\If{$t==1$}
\State $L^{t}=\mathcal{F}(\mathcal{T},~\mathcal{O},~\mathcal{C}^{t-1},~\mathcal{M};~\mathcal{P})$ \Comment{Get output logits $L^{t}$ (see Figure~\ref{fig:sample} for input illustration)}
\State $Q = {L^{t}[len(\mathcal{T})]}$ \Comment{$Q$ is output logits of the last token of $\mathcal{T}$, \emph{i.e.} an AR output}
\State $a \sim Q$ \Comment{Sample the top-1 prediction $a$ as the newly generated token}
\State $\mathcal{O}.append(a)$ \Comment{Append $a$ to $\mathcal{O}$}
\State Get $\mathcal{C}^{t}$ by selecting the top $k$ predictions for each mask tokens $m_i~(i\in[1,~n])$ \Comment{Get draft candidates ${C}^{t}$}
\Else
\State $L^{t}=\mathcal{F}(\mathcal{T},~\mathcal{O},~\mathcal{C}^{t-1},~\mathcal{M} \times (n \cdot k + 1);~\mathcal{P})$ \Comment{Get output logits $L^{t}$ with $(n \cdot k + 1)$ groups of mask tokens}
\State $l=len(\mathcal{T} + \mathcal{O})$
\State $Q = L^{t}[l]$
\State $a \sim Q$
\For{$i=1$ to $n$}
\For{$j=1$ to $k$}
\If{$a==c_{i,j}^{t-1}$}
\State $\mathcal{O}.append(a)$ \Comment{$c_{i,j}^{t-1}$ is accepted}
\State $Q = L^{t}[l + j]$ \Comment{$Q$ is output logits of $c_{i,j}^{t-1}$, \emph{i.e.} an AR output}
\State $a \sim Q$
\If{$j==1$}
\State $l = l + k$ \Comment{Accept the top-1 scoring candidate then continue}
\EndIf
\State break
\EndIf
\EndFor
\If{$j > 1$}
\State break \Comment{Only consider ``children'' of the top-1 scoring candidate in the token tree}
\EndIf
\EndFor
\State $\mathcal{O}.append(a)$
\State Get $\mathcal{C}^{t}$ by selecting the top $k$ predictions for each mask tokens $m_r~(r\in[1,~n])$ \par
\Comment{$m_r$ is selected from the mask token group linked to either the last accepted candidate token or the last token of the input sequence if no candidate token is accepted in this step.}
\EndIf
\State $t=t+1$
\If{$<$EOS$>$ in $\mathcal{O}$}
\State $\mathcal{O} = \mathcal{O}$[:$index_{EOS}$]
\State break
\EndIf
\EndWhile
\State Return $\mathcal{O}$
\end{algorithmic}
\end{algorithm}

\end{document}